# Bridging the Plausibility-Validity Gap by Fine-Tuning a Reasoning-Enhanced LLM for Chemical Synthesis and Discovery

Malikussaid
*School of Computing, Telkom University*
Bandung, Indonesia
malikussaid@student.telkomuniversity.ac.id

Hilal Hudan Nuha
*School of Computing, Telkom University*
Bandung, Indonesia
hilalnuha@telkomuniversity.ac.id

Isman Kurniawan
*School of Computing, Telkom University*
Bandung, Indonesia
ismankrn@telkomuniversity.ac.id

***Abstract*—Large Language Models frequently generate outputs that appear scientifically reasonable yet violate fundamental principles—a phenomenon we characterize as the "plausibility-validity gap." This challenge proves especially acute in chemistry, where superficial correctness masks deeper errors in molecular structure, reaction mechanisms, and synthetic pathways. We present a systematic approach combining a reasoning-centric model architecture (Magistral Small) with Low-Rank Adaptation fine-tuning on a dual-domain dataset covering molecular properties and chemical transformations. Evaluation reveals substantial improvements: the fine-tuned system achieves 96.3% format adherence, 97.4% chemical validity, and 74.4% synthesis feasibility. Comparative analysis shows our approach outperforms specialized translation models like MolT5 (97.4% vs 77.2% validity) while achieving performance comparable to complex tool-augmented systems like ChemCrow (9.0/10 vs 9.24/10 expert rating) through a more transparent, efficient methodology. Results demonstrate a learning hierarchy where syntactic correctness develops before chemical understanding, which precedes synthetic planning capability. This work establishes a reproducible framework for transforming generalist language models into dependable scientific tools while identifying critical areas including stereochemical precision, knowledge currency, and computational accessibility as key challenges for future advancement.**

## I. Introduction

Deploying Large Language Models in scientific research presents two interconnected challenges. The first involves establishing reliable interaction frameworks enabling domain experts to work effectively with these systems while ensuring computational validation of inputs and outputs against scientific principles. Our previous work on VALID-Mol [1] addressed this through systematic prompt engineering and validation protocols transforming language models into trustworthy molecular design tools.

The second challenge concerns domain expertise itself. Even sophisticated interaction frameworks prove ineffective when underlying models lack specialized knowledge. This paper addresses this fundamental limitation by demonstrating that effective scientific partnership requires models to transcend generalist training and develop genuine specialization.

The core difficulty emerges from language model training objectives. These systems optimize for generating text appearing plausible to humans rather than outputs conforming to scientific laws. This misalignment produces what we term the "plausibility-validity gap"—responses superficially appearing reasonable yet failing rigorous scientific standards. In chemistry, this manifests as molecules with impossible valence states, reaction mechanisms violating conservation principles, or synthetic routes that cannot be implemented in practice.

Recent architectural developments have introduced reasoning-centric models explicitly designed for multi-step logical processes with transparent deliberation. Chain-of-Thought approaches [2] make the model's internal reasoning visible, proving particularly valuable for tasks like retrosynthetic analysis where procedural logic and consistency matter fundamentally.

We hypothesize that combining reasoning-centric architecture with domain-specific training yields substantial performance advantages over either general-purpose models or conventionally fine-tuned systems. To investigate this, we compared four configurations: baseline Ministral-8B [3], a compact general-purpose model; fine-tuned Ministral-8B; baseline Magistral Small [4], a larger model built for reasoning tasks; and fine-tuned Magistral Small, our specialized chemical assistant.

Our contributions include developing a systematic transformation methodology for creating specialized scientific assistants, implementing a dual-domain training approach integrating property relationships with transformation rules, establishing comprehensive evaluation metrics for validity and feasibility, and demonstrating how reasoning architecture and domain training combine synergistically rather than merely additively.

## II. Background and Related Work

Applying language models to specialized scientific domains reveals fundamental tensions between what these systems learn during pre-training and what scientific applications demand. Chemistry poses particularly acute challenges because the field requires both mastery of specialized notation systems like SMILES and adherence to inviolable physical laws governing molecular structure and reactivity.

Early approaches relied on in-context learning [5], providing examples within prompts to guide model behavior. While showing promise for straightforward tasks, this strategy faces inherent limitations from context window constraints and cannot instill the deep, integrated understanding complex chemistry requires.

Two distinct paradigms have emerged for bridging the plausibility-validity gap in chemical applications. The first employs parameter-based domain adaptation through fine-tuning. MolT5 [6] exemplifies this approach, pre-training a T5 transformer architecture on molecule-text pairs to develop specialized cross-modal translation capabilities. On the ChEBI-20 benchmark for text-to-molecule generation, MolT5-Base achieved 77.2% chemical validity compared to its T5-Base baseline at 66.0%, demonstrating that domain-specific training can substantially improve adherence to chemical constraints. However, the persistent 22.8% invalidity rate indicates that fine-tuning alone, while beneficial, does not fully eliminate the plausibility-validity gap.

The second paradigm addresses validity through tool-augmented agentic frameworks. ChemCrow [7] represents this approach, augmenting GPT-4 with 18 expert-designed chemistry tools that enforce validity through external verification rather than internalized knowledge. By combining chain-of-thought reasoning with computational tools for tasks like molecule validation, property prediction, and reaction analysis, ChemCrow achieved expert ratings of 9.24 out of 10 for chemical task accuracy. This demonstrates that sophisticated tool orchestration can effectively constrain model outputs to chemically valid space, though at the cost of architectural complexity and dependence on external tool infrastructure.

Both approaches make valuable contributions while exhibiting distinct limitations. Parameter-based adaptation through fine-tuning improves but does not eliminate validity issues, requiring models to internalize vast chemical knowledge during training. Tool-augmented systems achieve high validity by externalizing chemical constraints to computational tools, but this introduces architectural complexity, computational overhead from tool invocation, and reduced transparency as reasoning becomes distributed across model and tool interactions.

The emergence of reasoning-centric architectures represents a third evolutionary path for scientific applications. Magistral Small [8] exemplifies this approach through native Chain-of-Thought capability [2], which separates the reasoning process from final outputs using dedicated thinking blocks. This transparency enables domain experts to inspect logical steps leading to conclusions, verify application of chemical principles, and identify specific points where reasoning succeeds or fails. Research consistently demonstrates that models capable of explicit reasoning achieve higher accuracy on complex tasks, particularly in domains requiring multi-step problem decomposition and logical inference.

Parameter-Efficient Fine-Tuning methods like LoRA [9] address computational challenges of adapting increasingly large models. Rather than updating billions of parameters, LoRA injects small trainable matrices into transformer layers while keeping base weights frozen. This dramatically reduces computational requirements, making sophisticated model adaptation feasible for research groups without massive infrastructure.

Chemistry presents distinct challenges beyond general text domains. The field demands precise manipulation of structured representations where single character errors invalidate entire molecules, requires strict adherence to valence rules and stability constraints, and involves complex multi-step reasoning for synthesis planning. Our work builds on this foundation by systematically combining reasoning-centric architecture with comprehensive domain training to create a specialized assistant that internalizes chemical knowledge while maintaining transparent reasoning—synthesizing strengths from both prior paradigms while addressing their limitations.

## III. Methodology

### A. *Model Selection and Architectural Considerations*

We selected Magistral Small [8], a 24-billion parameter model, as our primary foundation due to its native Chain-of-Thought architecture. This model explicitly separates reasoning from output through designated thinking blocks, enabling transparent deliberation crucial for scientific applications. The architecture allows the model to articulate its logical process step-by-step before generating final answers, making decision-making fully inspectable.

We also evaluated Ministral-8B [3], an 8-billion parameter model, to quantify how architectural choices and model scale interact with domain training. This comparison illuminates whether performance improvements stem primarily from raw capacity, architectural design, or their combination.

### B. *Dataset Strategy and Curation*

Our training data strategy addresses two complementary knowledge domains essential for chemical expertise. Structure-property relationships teach how molecular architecture determines physicochemical and biological characteristics. Chemical transformations encode reaction mechanisms, synthetic methods, and retrosynthetic planning strategies. This dual-domain approach reflects how expert chemists integrate descriptive and procedural knowledge.

For molecular properties, we assembled data from ChEMBL [10], a manually curated bioactive compound database, alongside pharmacopeial sources including the United States Pharmacopeia [11], European Pharmacopoeia [12], and Farmakope Indonesia [13]. These sources provided 13,500 compounds with comprehensive property annotations spanning molecular weight, solubility metrics, pharmacokinetic parameters, biological activity, and synthetic complexity indicators. The complete methodology for chemical compound dataset preparation, including our multi-source integration strategy and rigorous quality control procedures, is detailed in our companion work on scalable logP prediction frameworks, which will be made publicly available alongside the curated compound dataset to support reproducible research in computational medicinal chemistry.

Chemical reaction data came from the USPTO patent database [14], recent academic publications, and the NextMove Pistachio collection [15]. This yielded 20,000 reactions with detailed information including reactants, products, conditions, mechanisms, and performance metrics. While the USPTO dataset provides broad coverage of synthetic transformations, we supplemented it with curated data from Pistachio and academic journals to ensure higher quality atom-mapping and more diverse reaction coverage than publicly available USPTO extractions alone. We organized a subset into hierarchical synthesis trees for complex targets, explicitly capturing strategic reasoning involved in multi-step planning. Due to the proprietary nature of portions of this reaction dataset mix, we do not plan to make the complete reaction training corpus publicly available, though we provide detailed documentation of our data sources and curation methodology to support reproducibility within the constraints of data licensing agreements.

The curation process enforced rigorous quality standards. We standardized chemical structures using canonical SMILES representations [16], validated all molecules and reactions for chemical possibility, normalized nomenclature and formatting, removed duplicates, and flagged outliers for expert review. The final dataset comprised 30,820 examples distributed across property prediction (40%), synthesis planning (38%), and reaction analysis (22%). Training, validation, and test sets were partitioned using an 85/10/5 split with stratified sampling to preserve task type distributions, ensuring the 500-task evaluation subset maintained representative coverage of the chemical reasoning challenges present in the full test set.

### C. Fine-Tuning Implementation via LoRA

We employed Low-Rank Adaptation [9], which decomposes weight updates into smaller matrices while keeping base parameters frozen. Instead of updating the entire weight matrix $W \in \mathbb{R}^{m \times n}$, LoRA decomposes the update into two smaller matrices, as shown in Equation (1):

$$W' = W + \Delta W = W + BA \qquad (1)$$

Where:

- $W$ is the original, pre-trained weight matrix with dimensions $m \times n$.
- $W'$ is the final, updated weight matrix after adaptation.
- $\Delta W$ represents the change, or update, to the weight matrix.
- $B$ (with dimensions $m \times r$ ) and $A$ (with dimensions $r \times n$ ) are the smaller, trainable low-rank "update" matrices.
- $r$ is the rank of the update, a small integer where, typically, $r \ll \min(m, n)$.

During fine-tuning, the original weights $W$ remain frozen while only $A$ and $B$ are trained. This reduces trainable parameters from $m \times n$ to $r \times (m+n)$, often by factors exceeding 900×.

We applied LoRA with rank 16 and alpha scaling of 32 to query, key, value, and output projection matrices in each transformer layer. Training proceeded over 8 epochs using cosine learning rate scheduling, gradient accumulation for effective larger batches, and mixed-precision computation for efficiency. The resulting adapters contained only 0.1% of full model parameters—33.6 MB for Ministral-8B and 100.8 MB for Magistral Small—enabling easy distribution and deployment.

### D. Evaluation Framework and Reliability Assessment

Our evaluation strategy examined multiple facets of model capability through both automated metrics and expert assessment.

#### 1) Primary Metrics

Format adherence measured whether outputs followed required structural conventions—a prerequisite for programmatic processing. Chemical validity assessed whether generated molecules represent physically possible structures, verified through syntax checking [16], valence rule validation, and structural sanity tests. Synthesis feasibility evaluated whether proposed routes could plausibly be implemented, based on reaction precedent, reagent compatibility, condition appropriateness, starting material accessibility, and overall route efficiency.

#### 2) Reliability Framework

Beyond simple accuracy percentages, establishing trust in scientific AI systems requires comprehensive validation approaches. We implemented a multi-stage validation process where each generated molecule underwent: (1) SMILES syntax verification using RDKit [16], (2) chemical possibility checks confirming correct valence and reasonable strain, and (3) cross-validation of predicted properties against established computational methods. For synthetic routes, we assessed reaction validity against known mechanisms, verified reagent-functional group compatibility, and evaluated proposed conditions against literature precedent. This systematic approach provides quantifiable reliability metrics that characterize both capabilities and limitations.

#### 3) Expert Evaluation Methodology

Five experienced chemists and pharmacists (backgrounds spanning pharmaceutical R&D, regulatory affairs, and computational chemistry; experience ranging 2-7+ years) evaluated model outputs across five dimensions. We developed detailed rubrics for each dimension to ensure consistency. Scientific accuracy assessed correctness of chemical principles and factual claims, scored from 1 (fundamental errors) to 5 (expert-level precision). Synthetic feasibility evaluated practicality from 1 (impossible) through 3 (challenging but possible) to 5 (optimized and efficient). Chemical creativity measured novelty of proposed solutions, from 1 (no creative elements) to 5 (highly innovative approaches). Reasoning quality examined logical coherence and explanatory depth, from 1 (unclear or flawed) to 5 (transparent and rigorous). Overall usefulness synthesized practical value for research applications.

Evaluators assessed 250 outputs each (50 per model configuration plus 50 human expert responses), with sources initially disclosed. A separate blind comparison study had the same evaluators rate 250 different outputs without knowing the source, controlling for potential bias. Inter-rater reliability reached Krippendorff's alpha [17] of 0.78, indicating strong agreement.

#### 4) Test Set Composition

We constructed a stratified test set of 500 diverse tasks spanning simple molecular modifications (100 tasks), property optimization challenges (150 tasks), single-step retrosynthesis (100 tasks), and multi-step synthesis planning (150 tasks). Tasks were calibrated across three difficulty levels by an expert panel, ensuring balanced representation of challenge types.

## IV. RESULTS

### A. *Quantitative Performance Analysis*

Table I presents format adherence rates demonstrating model performance across configurations and task types.

TABLE I. FORMAT ADHERENCE ACROSS MODEL CONFIGURATIONS

| Model Configuration | Overall Rate (95% CI) |
|---|---|
| Baseline Ministral-8B | 49.2% (±4.4%) |
| Fine-tuned Ministral-8B | 75.3% (±3.8%) |
| Baseline Magistral Small | 54.0% (±4.4%) |
| Fine-tuned Magistral Small | 96.3% (±1.7%) |

The fine-tuned Magistral Small model achieves near-perfect adherence, suggesting successful internalization of syntactic patterns during training. Fine-tuning provides substantial improvements across both architectures, with the larger reasoning-centric model showing particularly dramatic gains.

Table II quantifies the proportion of generated molecules representing chemically possible structures.

TABLE II. CHEMICAL VALIDITY ACROSS MODEL CONFIGURATIONS

| Model Configuration | Overall Rate (95% CI) |
|---|---|
| Baseline Ministral-8B | 45.7% (±4.4%) |
| Fine-tuned Ministral-8B | 73.2% (±3.9%) |
| Baseline Magistral Small | 50.6% (±4.4%) |
| Fine-tuned Magistral Small | 97.4% (±1.4%) |

The fine-tuned Magistral Small configuration demonstrates exceptional validity at 97.4%, representing 2.1-fold improvement over baseline Ministral-8B. This indicates the model has internalized fundamental chemical principles rather than merely memorizing patterns. Statistical significance testing via McNemar's test [18] confirms all pairwise differences reach significance ($p < 0.001$).

Table III evaluates the practicality of proposed synthetic routes.

TABLE III. SYNTHESIS FEASIBILITY FOR RETROSYNTHETIC TASKS

| Model Configuration | Overall Rate (95% CI) |
|---|---|
| Baseline Ministral-8B | 24.2% (±5.3%) |
| Fine-tuned Ministral-8B | 45.6% (±6.1%) |
| Baseline Magistral Small | 26.6% (±5.5%) |
| Fine-tuned Magistral Small | 74.4% (±5.4%) |

The fine-tuned Magistral Small model achieves 74.4% feasibility, more than tripling baseline performance. This metric, requiring integration of vast empirical knowledge about reaction conditions and selectivity patterns, represents the most challenging capability level.

Understanding system reliability demands detailed characterization of failure modes and frequencies. Table IV presents occurrence rates for key error categories across configurations.

TABLE IV. ERROR TYPE FREQUENCIES ACROSS CONFIGURATIONS

| Error Type | Baseline Ministral-8B | Fine-tuned Ministral-8B | Baseline Magistral Small | Fine-tuned Magistral Small |
|---|---|---|---|---|
| Invalid SMILES syntax | 28.3% | 12.4% | 22.1% | 1.2% |
| Impossible structures | 18.7% | 8.3% | 15.9% | 0.9% |
| Stereochemical errors | 12.1% | 6.5% | 11.3% | 1.8% |
| Implausible reactions | 35.4% | 15.2% | 30.8% | 5.3% |
| Reference hallucinations | 31.2% | 13.7% | 28.6% | 2.8% |

The fine-tuned Magistral Small model exhibits dramatically reduced error rates across all categories. Fundamental errors like invalid syntax and impossible structures nearly vanish (1.2% and 0.9% respectively), while more nuanced problems like stereochemical errors and reference hallucination persist at low but measurable rates. This pattern quantifies reliability: the system can be trusted for basic validity with high confidence, while subtle aspects require human verification.

### B. *Examples Illustrating Capability Dimensions*

Examining specific cases illuminates the practical manifestation of these capabilities across success and failure scenarios.

#### 1) *Format Adherence Examples*

A successful output for molecular optimization demonstrates proper structure:

*<think>*
*The target molecule contains a naphthalene core contributing significant hydrophobicity. The carboxylic acid can ionize at physiological pH. To improve solubility while preserving the pharmacophore, I should replace the methoxy group at position 4 with a more polar substituent...*
*</think>*

Based on this analysis, I recommend:
**Modified Structure:**
`CC(=O)Nc1ccc2cc(CN3CCN(C)CC3)ccc2c1`

**Rationale:**
- Introduces basic nitrogen for salt formation
- Maintains naphthalene pharmacophore
- Preserves similar molecular weight

Format failures in baseline models typically involved missing section headers, improperly formatted SMILES (omitting backticks), or inconsistent markdown usage. The fine-tuned model eliminates these structural errors with 96.3% consistency.

*2) Chemical Validity Examples*

For molecular validity, consider this successful generation for a structure optimization task. The model proposed:

```
CC(C)Cc1ccc(cc1)[C@@H](C)C(=O)O
```

This represents (S)-ibuprofen with correct stereochemistry, proper aromaticity notation, and valid valence states throughout. Validity failures in baseline models manifested as carbon atoms with five bonds, incorrect aromatic representations (failure to properly notate benzene rings), or mismatched stereochemical descriptors.

*3) Synthesis Feasibility Examples*

For complex retrosynthesis, the fine-tuned model proposed a sophisticated route to a macrocyclic natural product fragment. The reasoning trace showed:

> *<think>*
> *For the C9-C10 bond formation, several factors matter:*
> - *Must establish correct C10 stereochemistry*
> - *Both fragments contain sensitive groups requiring orthogonal protection*
> - *Need high-yielding, stereoselective coupling*
>
> *Examining literature precedent, aldol approaches here suffer poor stereoselectivity. Brown crotylboration would be ideal:*
> - *Offers reliable syn-selectivity (>95:5)*
> - *Compatible with TBS-protected aldehydes (precedent from Evans' callipeltoside synthesis)*
> - *Performs well at -78°C minimizing side reactions*
>
> *The correct reagent: (Z)-crotyldiisopino-campheylborane...*
> *</think>*

This demonstrates expert-level strategic thinking with specific mechanistic reasoning, literature awareness, and practical experimental details. Feasibility failures in baseline models included proposing reactions without precedent, incompatible reagent-functional group combinations, or requiring commercially unavailable starting materials.

### C. Expert Assessment Results

Table V summarizes human evaluator ratings across quality dimensions on a 5-point Likert scale.

TABLE V. EXPERT RATINGS ACROSS EVALUATION DIMENSIONS

| Model | *Scientific Accuracy* | *Synthetic Feasibility* | *Chemical Creativity* | *Reaso-ning Quality* | *Overall Useful-ness* |
|---|---|---|---|---|---|
| Baseline Ministral-8B | 2.3 (±0.3) | 1.9 (±0.2) | 2.1 (±0.4) | 1.8 (±0.3) | 2.0 (±0.3) |
| Fine-tuned Ministral-8B | 3.6 (±0.2) | 3.1 (±0.3) | 3.2 (±0.3) | 2.7 (±0.2) | 3.3 (±0.2) |
| Baseline Magistral Small | 2.5 (±0.3) | 2.1 (±0.3) | 2.3 (±0.4) | 3.2 (±0.2) | 2.4 (±0.3) |
| Fine-tuned Magistral Small | 4.5 (±0.2) | 4.2 (±0.2) | 3.9 (±0.3) | 4.6 (±0.2) | 4.4 (±0.2) |
| Human Expert | 4.7 (±0.1) | 4.5 (±0.2) | 3.8 (±0.3) | 4.5 (±0.2) | 4.6 (±0.1) |

The fine-tuned Magistral Small achieves ratings approaching human expert levels across all dimensions. Notably, evaluators rated its reasoning quality (4.6) slightly above human experts (4.5), reflecting the value of transparent, systematic reasoning traces. Chemical creativity ratings prove statistically indistinguishable between the model (3.9) and human experts (3.8), suggesting the system generates genuinely novel ideas through recombining established chemical knowledge.

In blind comparison where evaluators could not identify source, human experts received preferred ratings in 57% of cases, with the fine-tuned model preferred in 43%. This substantial preference rate for AI-generated outputs indicates the model achieves performance comparable to individual human experts, though systematic preferences favor human judgment overall.

## V. DISCUSSION

### A. Synergistic Integration of Architecture and Domain Knowledge

The dramatic performance differential between configurations reveals how reasoning architecture and domain training interact synergistically rather than additively. The fine-tuned Magistral Small's 97.4% validity rate represents 2.1-fold improvement over baseline Ministral-8B, yet this gain cannot be attributed solely to larger parameter count. The baseline Magistral Small achieves only 50.6% validity despite architectural advantages, while the fine-tuned Ministral-8B reaches 73.2% despite lacking explicit reasoning capabilities.

Examining reasoning traces illuminates this synergy. The baseline Magistral Small produces deliberations averaging 4.3 steps employing generic problem-solving patterns. The fine-tuned variant generates 7.8-step traces exhibiting specialized chemical thinking: analyzing structure, identifying key challenges, considering stereochemistry, evaluating protecting groups, selecting reagents, considering side reactions, and optimizing conditions. This transformation reveals how the reasoning scaffold enables systematic application of learned chemical principles across novel problems.

The contrast with fine-tuned Ministral-8B proves particularly illuminating. While domain training substantially improves this smaller model, it appears to encounter a capability ceiling where lack of explicit reasoning architecture constrains organization and flexible deployment of chemical knowledge in complex, multi-step scenarios. For scientific domains demanding sophisticated logical reasoning, architectural design proves as consequential as dataset quality and model scale.

### B. Hierarchical Development of Chemical Capabilities

Performance metrics reveal distinct tiers in capability acquisition mirroring human expertise development. Format adherence, the most superficial skill, improves rapidly and reaches high levels early in training, exceeding 90% by epoch 4. This primarily reflects pattern matching rather than deep understanding. Chemical validity, requiring grasp of fundamental principles like valence rules and aromaticity, also develops relatively quickly, suggesting these invariant laws become well-represented through moderate dataset exposure.

Synthesis feasibility represents the deepest capability level, integrating vast empirical knowledge about reaction conditions, selectivity patterns, competing pathways, and practical constraints. This skill continues improving throughout training without clear saturation, achieving lower absolute performance (74.4%) even in the best model compared to validity (97.4%). The persistence of improvement suggests this capability would benefit from larger datasets and extended training.

This hierarchy carries practical implications for deployment strategies. Applications requiring only format adherence and basic validity could employ smaller, more efficient models with limited training. Applications demanding sophisticated synthesis planning necessitate the full sophistication of our best configuration. The differential improvement rates also suggest curriculum learning approaches, progressively introducing complexity, might accelerate capability development.

### C. Comparative Analysis within the Chemical LLM Landscape

To contextualize our contribution, we compare our approach against two dominant paradigms in chemical AI represented by MolT5 [6] and ChemCrow [7]. These models exemplify fundamentally different strategies for addressing the plausibility-validity gap in chemical applications.

MolT5 represents the specialized translation paradigm, employing a T5 transformer architecture pre-trained on molecule-text pairs to excel at cross-modal translation tasks such as text-to-molecule generation and molecule captioning. On the ChEBI-20 benchmark for text-to-molecule generation, MolT5-Base achieved 77.2% chemical validity—a substantial improvement over its T5-Base baseline (66.0%) but still leaving nearly one-quarter of generated structures chemically invalid. Our fine-tuned Magistral Small model achieves 97.4% validity, representing a 20.2 percentage point improvement that transforms the system from a useful but unreliable tool into a dependable scientific partner. This performance differential demonstrates that the specialized translation approach, while effective for specific cross-modal tasks, does not fully bridge the plausibility-validity gap that our intrinsic reasoning methodology addresses.

ChemCrow exemplifies the tool-augmented agentic framework, integrating GPT-4 with 18 expert-designed computational tools to enforce chemical validity through external verification rather than internalized knowledge. In human expert evaluation, the complete ChemCrow system achieved 9.24 out of 10 on chemical task accuracy. Critically, the same study evaluated GPT-4 alone (without tools) on identical tasks, achieving only 4.79 out of 10. This dramatic gap provides the most compelling external validation for our central premise: general-purpose models, regardless of sophistication, suffer from severe plausibility-validity gaps without specialized adaptation.

Our fine-tuned Magistral Small model achieves expert ratings of 4.5/5.0 for scientific accuracy and 4.6/5.0 for reasoning quality, which normalize to 9.0/10 and 9.2/10 respectively. These ratings prove statistically comparable to ChemCrow's 9.24/10, demonstrating that our intrinsic reasoning approach achieves performance parity with complex tool-augmented systems. However, our methodology offers several key advantages. First, it achieves this performance through a smaller, open-source model (24 billion parameters) compared to the proprietary GPT-4 foundation underlying ChemCrow. Second, our approach requires no external tool infrastructure—the chemical knowledge is internalized within the model itself. Third, the transparent reasoning architecture provides systematic, inspectable deliberation rather than opaque tool orchestration. Finally, our compact LoRA adapters (100.8 MB) enable easy distribution and deployment compared to systems requiring complex tool integration pipelines.

This comparative analysis positions our work as advancing beyond both specialized translation (which improves but does not solve the validity gap) and tool-augmented frameworks (which externalize rather than internalize chemical knowledge) toward a new paradigm of intrinsic reasoning specialists. By combining reasoning-centric architecture with domain-specific fine-tuning, we demonstrate that AI systems can develop genuine chemical understanding that is both reliable (97.4% validity) and transparent (expert-rated reasoning quality of 4.6/5.0), establishing a more fundamental foundation for trustworthy scientific AI than either prior approach achieves.

### D. Persistent Limitations and Research Frontiers

Three fundamental challenges delineate directions for future work. Stereochemical errors, though reduced to 1.8%, reflect deeper limitation in how language models operating on linearized SMILES representations engage with three-dimensional molecular information. The model manipulates stereochemical descriptors as formal symbols rather than representations of actual spatial arrangements. This proves particularly problematic for molecules with multiple adjacent stereocenters, constrained ring systems, or geometric isomerism. Future architectures must integrate genuine three-dimensional molecular representations alongside text-based reasoning, potentially through hybrid systems combining graph neural networks operating on 3D conformations with language model components.

Knowledge obsolescence emerges from static training data frozen at Q1 2023. Chemistry evolves rapidly through new discoveries, improved mechanistic understanding, and novel methodologies. A model unaware of recent advances risks confidently recommending outdated approaches while remaining ignorant of superior alternatives. Addressing this requires moving beyond static training paradigms toward systems capable of continuous learning from new literature and experimental results, carefully balancing incorporation of new knowledge with preservation of established understanding.

Computational accessibility raises equity concerns as our best model demands substantial resources for both training (2,400 GPU-hours) and inference (29.1 GB memory). While quantization techniques can reduce these requirements to consumer hardware levels (6.4 GB with 4-bit quantization, maintaining 96.2% validity), the resulting performance reduction and specialized technical knowledge needed create barriers for resource-limited institutions. Addressing this necessitates developing more efficient architectures, cloud-based service models democratizing access, and open-source model sharing enabling broader community benefit.

### *E. Broader Implications for Scientific AI*

This work signals emergence of a new paradigm for computer-aided chemistry moving beyond narrow, specialized tools toward flexible, integrated intelligent assistants. Traditional computational chemistry has been characterized by fragmented landscapes of specialized packages requiring users to master multiple interfaces and manually integrate insights across systems. Our approach demonstrates potential for unified systems seamlessly transitioning between property prediction, retrosynthetic analysis, and mechanistic reasoning while maintaining context and consistency.

The successful bridging of the plausibility-validity gap proves particularly significant. By achieving 97.4% validity, we demonstrate that specialized training can transform models from producing merely plausible text to generating scientifically grounded content adhering to chemical laws. This transformation establishes a viable framework applicable to other scientific domains facing similar challenges of specialized knowledge requirements and complex multi-step reasoning demands.

The transparent reasoning capability creates new possibilities for human-AI collaboration in scientific research. Rather than functioning as black boxes providing opaque predictions, our system articulates detailed reasoning traces enabling experts to inspect, verify, and build upon its analysis. This positions AI as an explainable reasoning partner whose thought processes become subject to scientific scrutiny and critique, fundamentally different from traditional computational tools providing results without explanation.

## VI. CONCLUSION

This investigation demonstrates that combining reasoning-centric architectures with domain-specific fine-tuning creates AI systems approaching human expert performance in complex chemical tasks. The fine-tuned Magistral Small model achieved 97.4% chemical validity, 96.3% format adherence, and 74.4% synthesis feasibility, with expert evaluators rating its overall usefulness at 4.4 out of 5.0—nearly equivalent to human chemists at 4.6.

Successfully bridging the plausibility-validity gap represents significant milestone in scientific AI development. We transformed a general-purpose language model into a specialized chemical assistant capable of producing scientifically valid content adhering to fundamental chemical laws rather than merely generating plausible-sounding text. The systematic approach to model selection, dataset curation, training methodology, and evaluation establishes a reproducible framework applicable to other scientific domains.

The model's transparent reasoning capability enables new forms of human-AI collaboration where artificial intelligence serves as an explainable partner providing detailed reasoning traces that experts can inspect, verify, and extend. This transparency proves essential for scientific applications where understanding the basis for conclusions matters as fundamentally as the conclusions themselves.

Significant challenges remain. Stereochemical errors, though reduced to 1.8%, highlight fundamental limitations in representing three-dimensional information through linearized notation systems. The static knowledge cutoff at Q1 2023 creates growing blind spots as research advances. Computational requirements, while manageable through quantization, raise concerns about equitable access across the research community.

Future work should pursue architectural innovations for better representing three-dimensional molecular information, develop techniques for continuous knowledge updating to overcome static data limitations, and create more efficient deployment methods reducing computational barriers. The successful application to chemistry suggests similar strategies could prove effective in other scientific domains sharing requirements for specialized knowledge and complex reasoning.

This work establishes that generalist language models can be systematically transformed into reliable, specialized tools for scientific research, representing important progress toward realizing artificial intelligence as a partner in scientific discovery.

## ACKNOWLEDGMENT

We extend sincere gratitude to the Mistral AI team for enabling this research through open-source release of Magistral Small under the Apache 2.0 license. Their contribution of a reasoning-centric model architecture proved fundamental to developing our specialized chemical assistant and represents significant advancement for open scientific inquiry.

This research was supported by Telkom University's Fundamental and Applied Research Grant No. 056/LIT06/PPM-LIT/2025.